%% file: conference_101719.tex
\documentclass[conference]{IEEEtran}
\IEEEoverridecommandlockouts
\usepackage{cite}
\usepackage[linesnumbered,ruled,vlined]{algorithm2e}
\usepackage{amsmath,amssymb,amsfonts}
\usepackage{algorithmic}
\usepackage{graphicx}
\usepackage{textcomp}
\usepackage[usenames,dvipsnames,table]{xcolor}
\usepackage{tikz}
\usepackage{pgfplots}
\usepackage{caption}
\usepackage{subcaption}
\usepackage{acronym}

\SetKwInput{KwInput}{Input}
\SetKwInput{KwOutput}{Output}

\def\BibTeX{{\rm B\kern-.05em{\sc i\kern-.025em b}\kern-.08em
    T\kern-.1667em\lower.7ex\hbox{E}\kern-.125emX}}

\newcommand{\eq}[1]{(\ref{#1})}

\newcommand{\bi}{\begin{itemize}}
\newcommand{\ei}{\end{itemize}}

\acrodef{ifm}[ifm]{\emph{Input Feature Map Size}}
\acrodef{ofm}[ofm]{\emph{Output Feature Map Size}}
\acrodef{ksize}[ksize]{\emph{Kernel Size}}

\acrodef{NN}[NN]{\emph{Neural Network}}
\acrodef{DNN}[DNN]{\emph{Deep Neural Network}}

\acrodef{MAC}[MAC]{\emph{Multiply–Accumulate Operation}}
\acrodef{CL}[CL]{\emph{Computational Load}}
\acrodef{CNN}[CNN]{\emph{Convolutional Neural Network}}
\begin{document}

\title{Energy Consumption of Neural Networks on NVIDIA Edge Boards: an Empirical Model\\
}

\author{   

\IEEEauthorblockN{Seyyidahmed Lahmer, Aria Khoshsirat, Michele Rossi, Andrea Zanella}
    
    \textit{Department of Information Engineering}\\
    \textit{University of Padova}\\
     Padova, Italy\\
    \{firstname.lastname\}@unipd.it
}

\maketitle

\begin{abstract}
Recently, there has been a trend of shifting the execution of deep learning inference tasks toward the edge of the network, closer to the user, to reduce latency and preserve data privacy. At the same time, growing interest is being devoted to the energetic sustainability of machine learning. At the intersection of these trends, in this paper we focus on the energetic characterization of machine learning at the edge, which is attracting increasing attention. Unfortunately, calculating the energy consumption of a given neural network during inference is complicated by the heterogeneity of the possible underlying hardware implementation. In this work, we aim at profiling the energetic consumption of inference tasks for some modern edge nodes by deriving simple but accurate models. To this end, we performed a large number of experiments to collect the energy consumption of fully connected and convolutional layers on two well-known edge boards by NVIDIA, namely, Jetson TX2 and Xavier. From these experimental measurements, we have then distilled a simple and practical model that can provide an estimate of the energy consumption of a certain inference task on these edge computers.  
We believe that this model can prove useful in many contexts as, for instance, to guide the search for efficient neural network architectures, as a heuristic in neural network pruning, to find energy-efficient offloading strategies in a split computing context, or to evaluate and compare the energy performance of deep neural network architectures.
\end{abstract}

\begin{IEEEkeywords}
Energy consumption, Deep Neural Networks, Edge Computing, Inference 
\end{IEEEkeywords}

\section{Introduction}

Machine learning is being used in many applications, exploiting the abundance of data in the modern era and delivering state-of-the-art performance on a huge number of tasks.
For new emerging mobile applications, the traditional way of running inference tasks in cloud computing facilities and sending back the predictions to the end users is not always feasible because of the need for preserving privacy and ensuring low latency. On the other hand, shifting the inference towards the end devices presents its challenges, due to the limited resources available on these devices. To tackle these limitations, computation offloading from resource-limited end users to more powerful edge servers is being advocated as a promising method to schedule and execute user-generated tasks~\cite{MR1}. 

In fact, {\it Edge Computing} not only can provide faster online computations, closer to the end users, but it can also exploit the smart distribution and scheduling of computations to benefit from renewable energy resources (RERs), so as to reduce the carbon footprint of computing technology~\cite{MR2}. Besides improving throughput and latency, the energy efficiency of edge networks has gained much attention lately. For example, reference \cite{MR3} studies the whole network’s energy consumption, including access points, edge servers, and user equipment for a computation offloading scenario. According to this paper, the more we push the computation from cloud servers to the network's edge, the more crucial it becomes to consider the energy consumption of the models that are being exploited by end user applications.

Although deep learning (DL)~\cite{DL} has been known for its great success in terms of accurate predictions in a wide variety of tasks, energy and memory requirements of modern DL architectures may make the use of large deep neural networks for edge computing challenging. Split computing techniques have been proposed to tackle this problem. They basically focus on splitting a neural network at different candidate points, and performing early exit at such candidate points to obtain a trade-off between computing effort and quality of the result. This facilitates the deployment of deep networks at the network's edge, see, e.g.,~\cite{SC}. Further, designing energy efficient neural networks that have the same prediction accuracy as their more power hungry versions is receiving much attention from the research community~\cite{quan1},\cite{survey}.

Overall, current developments are evolving along two main axes: (i) providing online and energy efficient schedulers for edge computing networks that allow end users to offload their tasks, e.g.,~\cite{MR1,MR2}, and (ii) devising new energy efficient DL architectures, also entailing but not limited to the split computing paradigm, e.g.,~\cite{SC,quan1,survey}. We advocate that proper designs along both axes would greatly benefit from accurate energy consumption models of DL, especially tailored to modern edge computing hardware. These models are largely missing in the literature and are the objective of the present work.

In most of the existing literature on edge task scheduling, the energy cost models that were used for predicting the energy consumption mainly used the number of CPU cycles required to perform the tasks~\cite{MR4} or the amount of workload that a task produces~\cite{MR5}, using simple equations that proportionally depend on the squared CPU frequency or on the workload. While these models were very valuable to derive initial theories and results on scheduling algorithms, they may not suit well with the parallelizable computations on modern multi-core processing unit architectures. In fact, an accurate energy consumption estimation tool requires one to take into account the architecture of the host device, the different parameters in the neural network model that can exploit the parallel hardware architectures, and the exact number of operations a neural network requires for inference.

In this paper, we propose an experimentally validated and simple energy consumption model for neural networks on recent NVIDIA Jetson edge computers. The model allows one to estimate the energy drained by performing inference tasks on DL models composed of fully connected and convolutional layers, without having to perform online measurements of the energy drained. As we elaborate in the following, the main indicator for the energy consumption is the total number of multiply and accumulate (MAC) operations that are performed, as expected. Based on this number, for a convolutional layer, the energy consumption shows a multi-modal behavior governed by the number of kernels that are exploited. The derived empirical model is fully described by two hardware dependent parameters, which are here provided for Jetson TX2 and Xavier NX boards from NVIDIA. The model fitting for a simpler fully connected layer follows a similar rationale, but only requires a single parameter and shows a single slope in the MAC {\it vs} energy plot. 

The remainder of this paper is organized as follows: The related work is briefly commented in Section~\ref{sec:related_work}. In Section~\ref{sec:methodology} we present the experiment setup and configurations. The observations and discussions, in addition to an energy estimation model are provided in Section~\ref{sec:energy_charcterization}. Finally, conclusions and future research lines are discussed in Section~\ref{sec:conc}.

\section{Related work}
\label{sec:related_work}

Profiling the power/energy consumption of running \ac{NN}s on low-power edge devices has gained an increasing attention in recent years. In \cite{nano}, the authors measured the power consumption of an entire \ac{NN} as well as single NN layers on an NVIDIA Jetson Nano. A framework that predicts the energy consumption of CNNs on the Jetson TX1 based on real measurements has been proposed in~\cite{tx1}. This work is however still very preliminary, as it just presents the general measurement setup/methodology and some limited results. For the Jetson TX2 device, in~\cite{rasptx2} the authors have reported the power consumption of GPU and CPU, the memory usage and the time of executing the test phase on a fixed small \ac{CNN} architecture. Although the results in this paper are measured from real hardware, no analytical model is provided to gauge the energy consumption of the edge board from the neural network parameters.

In a research paper more similar to our present work, but based on simulations instead of real measurements~\cite{8335698}, the authors have provided an energy estimation tool for different types of neural network layers. They have shown that the energy consumption is not always proportional to the number of computations or parameters involved in a layer. Our results somehow confirm these observations, since the pure number of operations, \textit{per se}, is not sufficient to characterize the energy consumption of the boards. Nonetheless, with a careful and systematic analysis of the collected measurements, we were able to identify the effect of the different computational model parameters on the energy consumption of a single inference stage and, hence, define a model that captures reasonably well the experimental behavior of the computing boards. 

To the best of our knowledge, this is the first work to explore the real-world effect of choosing different configurations of a NN layer on the energy consumption of two NVIDIA Jetson edge devices (TX2 and Xavier NX), providing a parameterized analytical energy estimation model based on empirical measurements. Our model allows estimating the energy consumption of any custom set of layer configurations in common feed-forward deep neural networks.

\section{Experimental setup}
\label{sec:methodology}
We experimentally characterize the energy consumption of two energy-efficient embedded computing devices from NVIDIA, namely, Jetson TX2, and Jetson Xavier NX. These two edge computers are currently being used in several fields such as manufacturing, agriculture, retail, life sciences, etc. For instance, an image processing algorithm for thermal events has been recently proposed for the Jetson TX2~\cite{thermalTx2}. The configurations of both devices are shown in Table \ref{tab:jet-tx2config} (Jetson TX2) and \ref{tab:jet-xavierconfig} (Jetson Xavier NX).

\begin{table}[h]
\caption{\label{tab:jet-tx2config}NVIDIA \textbf{Jetson TX2} configurations}
\centering
\begin{tabular}{| p{0.2\columnwidth} p{0.7\columnwidth} |} 
  \hline
  \textbf{CPU}& Quad-Core ARM Cortex-A57 @ 2~GHz + Dual-Core NVIDIA Denver2 @ 2~GHz  \\ 
  \hline
  \textbf{GPU}& NVIDIA  Pascal 256 CUDA  cores @ 1300~MHz  \\ 
  \hline
  \textbf{Memory} & 8~GB 128-bit LPDDR4 @ 1866~Mhz, 59.7~GB/s \\ 
  \hline
  \textbf{Performance} & 1.3~TFLOPS \\ 
  \hline
\end{tabular}
\end{table}


\begin{table}[h]
\caption{\label{tab:jet-xavierconfig}NVIDIA \textbf{Jetson Xavier NX} configurations}
\centering
\begin{tabular}{| p{0.2\columnwidth} p{0.7\columnwidth} |} 
  \hline
  \textbf{CPU}&6-core NVIDIA Carmel ARM®v8.2 64-bit CPU 6~MB L2 + 4~MB L3  \\ 
  \hline
  \textbf{GPU}& 384-core NVIDIA Volta™ GPU with 48 Tensor Cores  \\ 
  \hline
  \textbf{Memory} & 8~GB 128-bit LPDDR4x 59.7~GB/s \\ 
  \hline
  \textbf{Performance} & 21~TFLOPS \\ 
  \hline
\end{tabular}
\end{table}

To assess the energy profile of these edge computers, we measure the timing and energy figures of neural network architectures, focusing on one single layer of the whole NN architecture. In fact, as demonstrated in~\cite{8335698}, and also independently verified by us, the energy consumption of two neural network layers $L_1$, $L_2$ that are executed in  sequence adds up, i.e., if their energy consumption is respectively $E(L_{1}) = E_{1}$ and $E(L_{2}) = E_{2}$, then sequentially using these two layers into a single model results in a total energy consumption of $E(L_{1}, L_{2}) \simeq E_{1} + E_{2}$, where the approximation accounts for the measurement noise and the intrinsic variability of the energy consumption of each single layer (as it will be seen later on in this paper). We hence focus our analysis on two widely utilized layer types, namely {\it fully connected} and {\it convolutional}, as better described in the following. 

\textbf{Fully Connected layer.} A fully connected layer consists of a bipartite set of input and output neurons, with each input neuron being connected to all the output neurons through weighted links. The output neurons apply a \mbox{non-linear} transformation to the weighted sum of the input vector, producing the corresponding output value. The following variables are hence used to describe a fully connected layer: 
\bi
\item {\rm i\_size}: Input feature map size, i.e., number of input neurons;
\item {\rm o\_size}: Output feature maps size, i.e., number of output neurons. 
\ei

We refer to the \textit{Computational Load} of a fully connected layer ($\rm CLF$) $L_i$ as the product of the number of input and output features of the layer, i.e., the value 
\begin{align}
    \label{eq:macs_count_fc}
    {\rm CLF}(L_{i}) & = \rm i\_size \times \rm o\_size.
\end{align}
Note, that ${\rm CLF}$ corresponds to the number of elements in the weight matrix and, hence, is proportional to the number of multiplications and additions that are performed as the layer is executed, i.e., to propagate the input to the output section.

\textbf{Convolutional layer.} We consider a generic convolutional layer defined by a multidimensional matrix of input values, with size $w \times h \times d$, and a set of kernel functions, each defined by a square matrix of real values of size $k \times k \times d$. Here, $d$ is referred to as the {\it depth} and should match the depth of the input feature map. 
Each kernel shifts along the input matrix with a step defined by another parameter called \textit{stride}. For each position, the dot product between the kernel and the corresponding elements of the input matrix is computed, and the results are then summed together to return one point of the output matrix. Each kernel generates one output map.

The following variables are then used to describe a convolutional layer: 
\bi
\item {\rm i\_size}: Input feature map size (i.e., $w = h$);
\item {\rm ifm}: Number of input feature maps (i.e., $d$); 
\item {\rm ofm}: Number of output feature maps (i.e., the number of kernel functions), 
\item  {\rm ksize} Kernel size parameter (i.e., $k$),
\item  {\rm stride}: Stride parameter (i.e., the sliding step of the kernel over the input matrix).  
\ei

We define the \textit{Computational Load} involved for a \textbf{single kernel} (KCLC) via the number of multiply-add operations, also referred to as Multiply–accumulate (MAC) operations, required to compute the convolution of the input maps with a single kernel (i.e., to obtain each one of the output maps). For a two dimensional convolutional layer $L_i$, neglecting the padding, it is obtained as follows
\begin{align}
    \label{eq:macs_count_c_k}
    {\rm KCLC}(L_{i}) & = \left (\dfrac{{\rm i\_size}_{i} - {\rm ksize}_{i}}{{\rm stride}_{i}} + 1 \right )^{2} \cdot {\rm ifm}_{i} \cdot {\rm ksize}^{2}_{i} \nonumber \\
    & \simeq \left (\dfrac{{\rm i\_size}_{i}}{{\rm stride}_{i}} \right )^{2} \cdot {\rm ifm}_{i} \cdot {\rm ksize}^{2}_{i},
\end{align}
where the approximation follows when the input size is much larger than the kernel size, which is typical in most practical cases. We then define the Computational Load for the whole convolutional layer (CLC) as Eq.~\eq{eq:macs_count_c}, 
\begin{align}
    \label{eq:macs_count_c}
    \rm CLC(L_{i}) & = {\rm KCLC}(L_{i}) \cdot \rm ofm_{i}.
\end{align}

The variables mentioned above are varied to generate different layers' configurations for the experiments. For each configuration, $50$ inference runs are performed using randomly generated non-zero inputs. Moreover, for each inference operation, the time is split into timeslots of the same duration $\delta=0.1$~ms and the power of the edge board is obtained from the onboard sensors at the end of each timeslot. The average energy consumed by the board over a time period of $T$ seconds is estimated as,
\begin{equation}
    \label{eq:board-energy-measure}
    \rm E_{\rm board}(T) \simeq T \cdot {\rm AverageBoardPower}(T).
\end{equation}

\section{Energy Characterization}
\label{sec:energy_charcterization} 

\subsection{Power consumption}
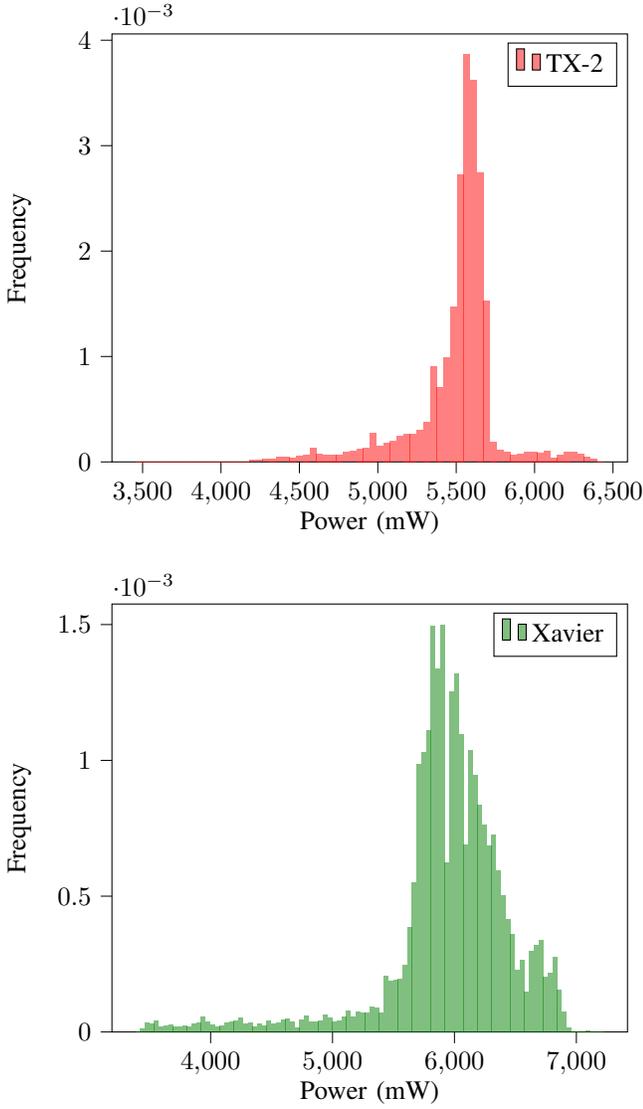
\begin{figure}[tbp]
    \begin{minipage}[t]{.45\textwidth}
        \centering
        \input{IEEEtran5/power-distribution-tx2}
        \label{fig:power-distribution-tx2}
    \end{minipage}
    \hfill
    \begin{minipage}[t]{.45\textwidth}
        \centering
        \input{IEEEtran5/power-distribution-xavier}
        \label{fig:power-distribution-xavier}
    \end{minipage}

    \caption{Distribution of the average board power consumption on Jetson TX 2 and Xavier NX for inference of many different neural network configurations.}
    \label{fig:power-distribution}
\end{figure}

Fig.~\ref{fig:power-distribution} reports the empirical histograms of the power consumption of the two boards when performing the inference tasks with different configurations of the convolutional or fully connected layers parameters.
As shown in these two figures, the power consumption follows a normal distribution for the different inference tasks, with the mean placed close to the amount of power consumption of the device when the CPU is at $100$~percent workload. 
This observation of the frequency of power consumption is especially relevant when the inference devices use renewable power sources, such as solar panels, that cannot provide more than a specific peak or mean power for a long period of time (due to the intermittent nature of renewable energy).

\subsection{Energy consumption when varying the model parameters}

\begin{figure*}[!ht]
    \begin{subfigure}{0.48\linewidth}
        \input{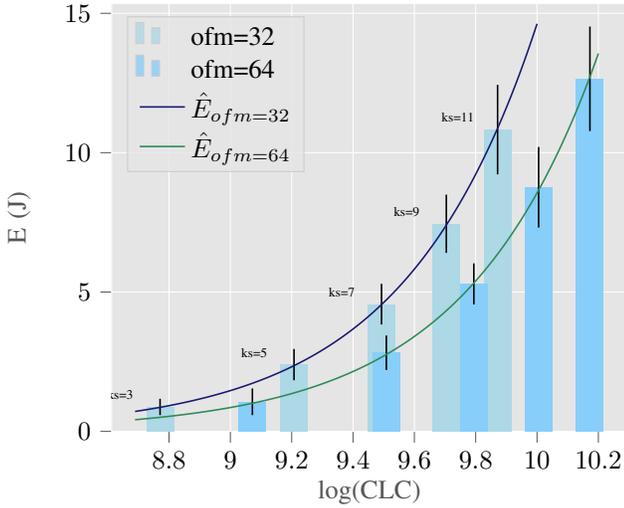}
        \caption{Varying the kernel size value for two different OFM values}
        \label{fig:a}
    \end{subfigure}
    \hfill
    \begin{subfigure}{0.48\linewidth}
        \input{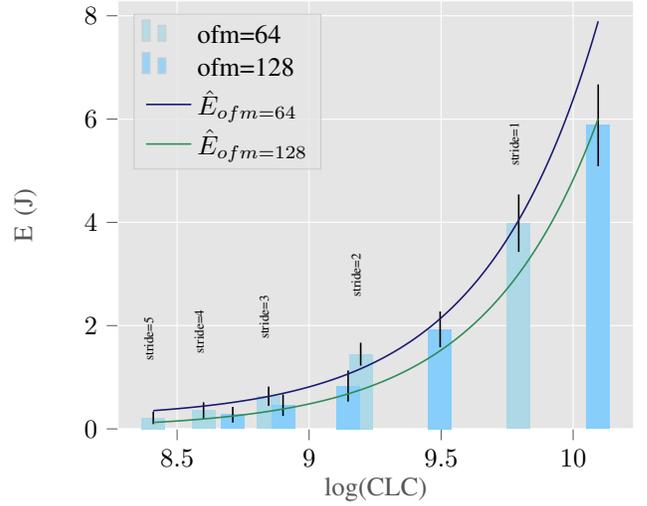}

        \caption{Varying the stride value for two different OFM values}
        \label{fig:b}
    \end{subfigure}
    \begin{subfigure}{0.48\linewidth}
        \input{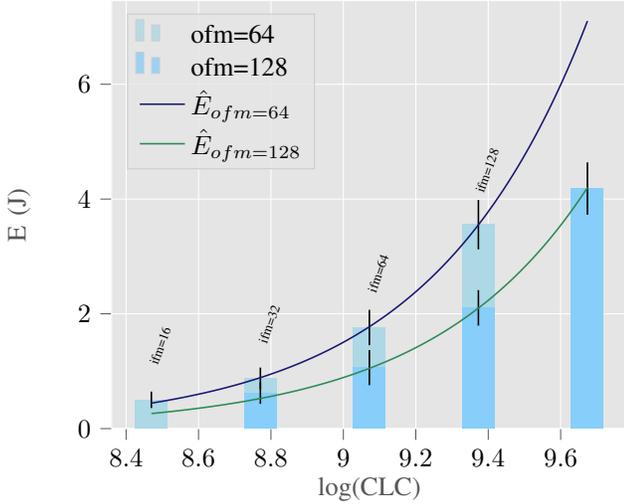}
        \caption{Varying the ifm value for two different OFM values}
        \label{fig:c}
    \end{subfigure}
    \hfill
    \begin{subfigure}{0.48\linewidth}
        \input{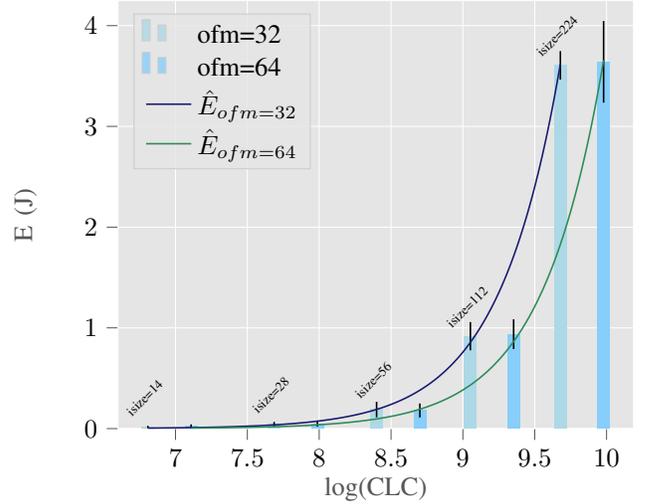}
        \caption{Varying the i\_size value for two different OFM values}
        \label{fig:d}
    \end{subfigure}
    
    \caption{Convolutional layer: exemplary demonstration of the linear relationship between CLC and the stride, ksize, ifm, and i\_size parameters; vertical bars show $99$\% confidence intervals. Continuous lines are obtained from our approximated model to gauge the average energy as a function of CLC ($\hat{E}$). The x-axis is represented in a $\log10$ scale.}
    \label{fig:all}
\end{figure*}

\begin{figure}[tbp]
\centering
\includegraphics[width=3.5in]{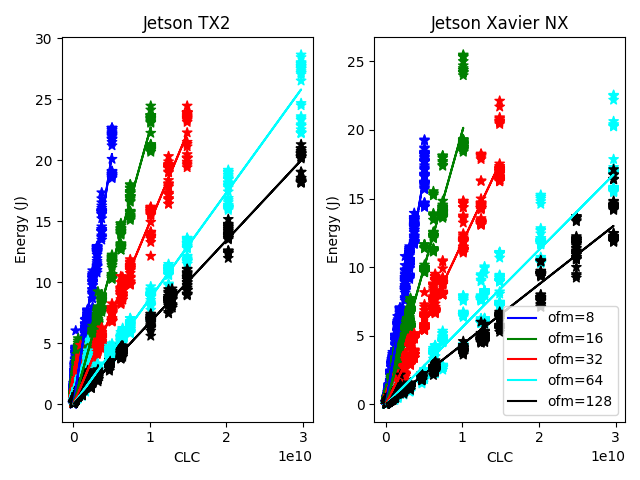}
\caption{The relationship between the computational load and the average energy growth for convolutional layer for a different values of ofm.}
\label{fig:energy-mac-tx2-xavier}
\end{figure}

\begin{figure}[tbp]
\centering
\includegraphics[width=3.5in]{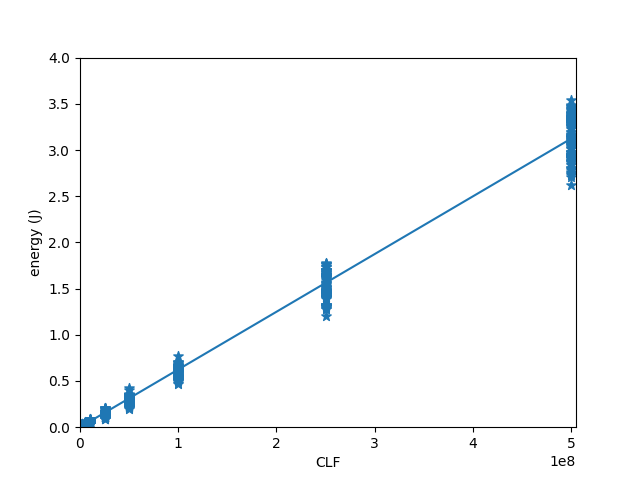}
\caption{The relationship between the average energy consumption and the computational load for a fully connected layer.}
\label{fig:energy-is-os-fc}
\end{figure}

Experimentally, we can understand the impact of the different parameters of each layer's configuration on the energy consumption; we came up with the following key observations:

\begin{itemize}
\item As it will be further analyzed in the next Section~\ref{subsec:energy_modeling}, the average energy drained for a convolutional layer $L_i$ is accurately approximated by $E_{\rm conv2d}(L_i) \simeq {\rm CLC}(L_i) \times H({\rm ofm}_i)$, where $H(\cdot)$ is a non-linear function (to be specified shortly).  
\item For a convolutional layer, the average energy consumption grows linearly with respect to CLC when varying the i\_size, ksize, ifm, or the stride. Fig.~\ref{fig:all} depicts this behavior: from the Eq.~\eq{eq:macs_count_c}, the CLC decreases polynomially with respect to an increasing stride, and so does the average energy; the CLC increases polynomially with increasing ksize and i\_size, and so does the average energy. The CL increases linearly with an increasing ifm, and so does the average energy.
\item Moreover, still for a convolutional layer $L_i$, the average energy consumption $E_{\rm conv2d}(L_i)$ does not grow linearly with CLC when changing the ofm parameter. From Fig.~\ref{fig:energy-mac-tx2-xavier}, and with both edge computing boards, we notice that the relationship between average energy and CLC can be interpolated through a linear function for a fixed ofm value. Also, we observe that the slope of this linear approximation changes (decreases) by changing the ofm (increasing). Function $H({\rm ofm})$ is here introduced to model this change of slope; it is also remarked that there is a noise term in the measurements, whose variance increases with an increasing CLC parameter.
\item For a fully connected layer, the average energy consumption grows linearly as a function of CLF (constant slope model), see Fig.~\ref{fig:energy-is-os-fc}.
\end{itemize}

\subsection{Energy modeling}
\label{subsec:energy_modeling}
\begin{figure}[tbp]
\centering
\includegraphics[width=3.5in]{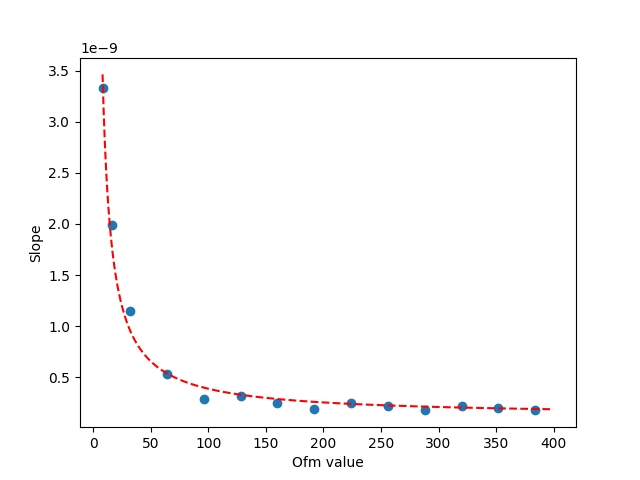}
\caption{The behaviour slope changes with respect to the ofm value represented by the blue points, with the corresponding interpolation represented by the red function.}
\label{fig:slope-ofm-tx2-xavier}
\end{figure}

To find a suitable shape for $H({\rm ofm})$, which maps an {\rm ofm} value onto a slope in the \mbox{MAC-vs-energy} plane, we generated additional data. The set of $N$ generated data pairs is denoted by $(x_{i},y_{i})$ where $i =1,\dots,N$, $x_{i}$ defines an {\rm ofm} value, and $y_{i}$ defines the corresponding slope $H(x_{i})$; Fig.~\ref{fig:slope-ofm-tx2-xavier} shows the function $H(\cdot)$, which approximately takes the form 
\begin{equation}
H({\rm ofm}_i) \simeq a_{c} \times \dfrac{1}{{\rm ofm}_i} + b_{c} .
\end{equation}
With this dataset, we use Mean Square Error (MSE) minimization as described in Eq.~\eq{eq:mse_proc} to estimate the parameters $a_{c}$, and $b_{c}$, which results in the red fitting curve in Fig.~\ref{fig:slope-ofm-tx2-xavier}.

\begin{align}
\label{eq:mse_proc}
{\rm MSE} & =  \dfrac{1}{N} \sum_{i=0}^{N-1} (H(x_{i}) - y_{i})^{2} 
\nonumber \\
& = \dfrac{1}{N} \sum_{\rm i=0}^{N-1} \left (a_{c} \times \dfrac{1}{x_{i}} + b_{c} - y_{i} \right )^{2} \nonumber \\
\frac{\partial {\rm MSE}}{\partial a_{c}} = 0 & \iff \frac{\partial \sum_{i=0}^{N-1} \left (a_{c} \times \dfrac{1}{x_{i}} + b_{c} - y_{i} \right)^{2}}{\partial a_{c}} & \!\!\!\!\!\!\!\!\!\! = 0 
\nonumber\\
& \iff a_{c} \sum_{i=0}^{N-1} \dfrac{1}{x_{i}^{2}} + b_{c} \sum_{i=0}^{N-1} \dfrac{1}{x_{i}} = \sum_{i=0}^{N-1} \dfrac{y_{i}}{x_{i}} 
\nonumber \\
\frac{\partial \rm MSE}{\partial b_{c}} & = 0  \iff b_{c} = \dfrac{1}{N} \left [ \sum_{i=0}^{N-1} y_{i} - a_{c} \sum_{i=0}^{N-1} \dfrac{1}{x_{i}} \right ]
\nonumber \\
a_{c} &= \dfrac{\sum_{i=0}^{N-1} \dfrac{y_{i}}{x_{i}} - \dfrac{1}{N} \sum_{i=0}^{N-1} \dfrac{1}{x_{i}} \sum_{i=0}^{N-1} y_{i}  }
{ \sum_{i=0}^{N-1} \dfrac{1}{x_{i}^{2}} - \dfrac{1}{N} \left (\sum_{i=0}^{N-1} \dfrac{1}{x_{i}} \right)^{2} }.
\end{align} 

With the previous key results and observations, we define the following model describing the average energy consumption for a convolutional layer, via Eq.~\eq{eq:energy_model}:

\begin{align}
    \label{eq:energy_model}
    E_{\rm conv2d}(L_i) & = {\rm CLC}(L_i) \times H({\rm ofm}_i)  \nonumber \\
     & = {\rm KCLC}(L_i) \times {\rm ofm}_{i} \times \left ( \dfrac{a_{c}}{{\rm ofm}_i} + b_{c} \right ) \nonumber \\
     & = {\rm KCLC}(L_i) \times (a_{c} + b_{c} \times {\rm ofm}_i).
\end{align}

For a fully-connected layer, we describe the average energy consumption through Eq.~\eq{eq:energy_model_fc} here below. The same procedure is followed to obtain the slope parameter $a_{f}$, with $x_{i}$ denoting the CLF, and $y_{i}$ denoting the corresponding average energy drained. The $b_{f}$ coefficient is set to zero, as with a zero CLF there is no energy expenditure.

\begin{align}
    \label{eq:energy_model_fc}
    E_{\rm fc}(L_{i}) & = {\rm CLF}(L_{i}) \times a_{f}.
\end{align}

Given a general CNN architecture description, the average energy expenditure of Conv2D and FC layers is gauged through the CLF and the platform-specific parameters $(a_{c},b_{c})$ (Conv2D layers) and $(a_{f})$ (FC layers), which we obtained empirically (see Table. \ref{tab:params}). 

Furthermore, for a feed-forward NN architecture with $\rm L$ layers, one can estimate the average energy consumption of the whole NN on the considered edge computing boards via Algorithm~1. 

\begin{table}[h]
\caption{\label{tab:params}Empirical Parameters}
\centering
\begin{tabular}{| l | c |}
  \hline
  \textbf{Parameter}  & \textbf{Value}  \\
  \hline
  \textbf{$a_{c}$ (TX2)}& 2.6727e-08 \\ 
  \hline
  \textbf{$b_{c}$ (TX2)}& 1.21334e-10  \\ 
  \hline
  \textbf{$a_{c}$ (Xavier NX)}& 2.8674e-08 \\ 
  \hline
  \textbf{$b_{c}$ (Xavier NX)}& 4.7639e-10  \\ 
  \hline
  \textbf{$a_{f}$ (Xavier NX)} & 6.2454e-09 \\
  \hline
\end{tabular}
\end{table}

\begin{algorithm}[h]
\DontPrintSemicolon
  \KwInput{\\
  $\rm nnArch = [(t,i/o\_size_{i}, ifm_{i},ofm_{i},ksize_{i},stride_{i})]^{L-1}_{i=0}$\\
  $\rm a_{conv},b_{conv},a_{fc} : Float$\tcc{Platform specific parameters}
  }
  \KwOutput{TotalEnergy}
  $\rm TotalEnergy = 0$
  
  \For{$\rm i \in [0..L-1]$}
   {
        \If{$\rm nnArch[i].t$ is CONV2D}
        {
          $\rm TotalEnergy = TotalEnergy + E_{conv2d}(nnArch[i], a_{c},b_{c})$
        }
        
        \ElseIf{$\rm nnArch[i].t$ is FC}
        {
          $\rm TotalEnergy = TotalEnergy + E_{fc}(nnArch[i], a_{f})$
        }
   }

\caption{Energy Estimator}
\end{algorithm}

\section{Concluding Remarks}
\label{sec:conc}
In this work, we have analyzed the energy consumption of \ac{NN}s on two NVIDIA edge boards, based on readings from the power sensors included in these devices. We have also investigated the effect of different parameters of convolutional and fully-connected layers on energy consumption during inference on CPU. Moreover, we have observed that the boards' peak and average power requirement when using CPU is less than when using GPU. This makes doing inference on CPU more inviting for limited power setups.

The energy estimation model provided in this work, which is backed-up with actual energy measurements on the edge devices, can help understand the effect of parameter choices on energy consumption for efficient development of new neural network architectures. It can also be used as a metric to optimise the scheduling of tasks on the network's edge, when energy efficiency is an important consideration.

In our future work, we plan to extend the experiments to additional edge devices and to other power profile settings of the edge boards, to study how inference can be customized in respect to power, latency and energy. Other interesting research directions include the investigation of effect of NN parameters when inference is performed on GPU and measuring the energy consumption for other layer types, e.g., long-short term memory (LSTM).

\section*{Acknowledgment}
This work has been supported by the EU H2020 MSCA ITN project Greenedge (grant no. 953775) and by MIUR (Italian Ministry for Education and Research) under the ”Departments of Excellence” initiative (Law 232/2016).

\bibliographystyle{IEEEtran}
\bibliography{Ref}

\end{document}

%% file: IEEEtran5/power-distribution-tx2.tex
\begin{tikzpicture}

\definecolor{darkgray176}{RGB}{176,176,176}

\begin{axis}[
tick align=outside,
tick pos=left,
x grid style={darkgray176},
xlabel={Power (mW)},
xmin=3304.55, xmax=6592.45,
xtick style={color=black},
y grid style={darkgray176},
ylabel={Frequency},
ymin=0, ymax=0.00405946902112154,
ytick style={color=black}
]
\draw[draw=none,fill=red,fill opacity=0.5] (axis cs:3454,0) rectangle (axis cs:3496.7,6.42734784694566e-07);
\addlegendimage{ybar,ybar legend,draw=none,fill=red,fill opacity=0.5}
\addlegendentry{TX-2}

\draw[draw=none,fill=red,fill opacity=0.5] (axis cs:3496.7,0) rectangle (axis cs:3539.4,0);
\draw[draw=none,fill=red,fill opacity=0.5] (axis cs:3539.4,0) rectangle (axis cs:3582.1,3.8316881395253e-07);
\draw[draw=none,fill=red,fill opacity=0.5] (axis cs:3582.1,0) rectangle (axis cs:3624.8,1.48323411852591e-07);
\draw[draw=none,fill=red,fill opacity=0.5] (axis cs:3624.8,0) rectangle (axis cs:3667.5,2.10124833457839e-07);
\draw[draw=none,fill=red,fill opacity=0.5] (axis cs:3667.5,0) rectangle (axis cs:3710.2,6.18014216052468e-08);
\draw[draw=none,fill=red,fill opacity=0.5] (axis cs:3710.2,0) rectangle (axis cs:3752.9,4.6969080419987e-07);
\draw[draw=none,fill=red,fill opacity=0.5] (axis cs:3752.9,0) rectangle (axis cs:3795.6,7.16896490620862e-07);
\draw[draw=none,fill=red,fill opacity=0.5] (axis cs:3795.6,0) rectangle (axis cs:3838.3,7.04536206299805e-07);
\draw[draw=none,fill=red,fill opacity=0.5] (axis cs:3838.3,0) rectangle (axis cs:3881,2.96646823705184e-07);
\draw[draw=none,fill=red,fill opacity=0.5] (axis cs:3881,0) rectangle (axis cs:3923.7,3.09007108026234e-07);
\draw[draw=none,fill=red,fill opacity=0.5] (axis cs:3923.7,0) rectangle (axis cs:3966.4,3.2136739234728e-07);
\draw[draw=none,fill=red,fill opacity=0.5] (axis cs:3966.4,0) rectangle (axis cs:4009.1,1.42143269692068e-06);
\draw[draw=none,fill=red,fill opacity=0.5] (axis cs:4009.1,0) rectangle (axis cs:4051.8,1.64391781469955e-06);
\draw[draw=none,fill=red,fill opacity=0.5] (axis cs:4051.8,0) rectangle (axis cs:4094.5,1.59447667741537e-06);
\draw[draw=none,fill=red,fill opacity=0.5] (axis cs:4094.5,0) rectangle (axis cs:4137.2,1.66863838334166e-06);
\draw[draw=none,fill=red,fill opacity=0.5] (axis cs:4137.2,0) rectangle (axis cs:4179.9,3.31255619804123e-06);
\draw[draw=none,fill=red,fill opacity=0.5] (axis cs:4179.9,0) rectangle (axis cs:4222.6,1.66369426961321e-05);
\draw[draw=none,fill=red,fill opacity=0.5] (axis cs:4222.6,0) rectangle (axis cs:4265.3,1.78358902752742e-05);
\draw[draw=none,fill=red,fill opacity=0.5] (axis cs:4265.3,0) rectangle (axis cs:4308,2.74027503397664e-05);
\draw[draw=none,fill=red,fill opacity=0.5] (axis cs:4308,0) rectangle (axis cs:4350.7,3.03444980081762e-05);
\draw[draw=none,fill=red,fill opacity=0.5] (axis cs:4350.7,0) rectangle (axis cs:4393.4,4.47071483892355e-05);
\draw[draw=none,fill=red,fill opacity=0.5] (axis cs:4393.4,0) rectangle (axis cs:4436.1,4.37554064965138e-05);
\draw[draw=none,fill=red,fill opacity=0.5] (axis cs:4436.1,0) rectangle (axis cs:4478.8,3.78719111596952e-05);
\draw[draw=none,fill=red,fill opacity=0.5] (axis cs:4478.8,0) rectangle (axis cs:4521.5,5.39526410613804e-05);
\draw[draw=none,fill=red,fill opacity=0.5] (axis cs:4521.5,0) rectangle (axis cs:4564.2,7.02187752278814e-05);
\draw[draw=none,fill=red,fill opacity=0.5] (axis cs:4564.2,0) rectangle (axis cs:4606.9,0.00012944925769435);
\draw[draw=none,fill=red,fill opacity=0.5] (axis cs:4606.9,0) rectangle (axis cs:4649.6,7.54966166329678e-05);
\draw[draw=none,fill=red,fill opacity=0.5] (axis cs:4649.6,0) rectangle (axis cs:4692.3,6.76478360891031e-05);
\draw[draw=none,fill=red,fill opacity=0.5] (axis cs:4692.3,0) rectangle (axis cs:4735,6.90569085017027e-05);
\draw[draw=none,fill=red,fill opacity=0.5] (axis cs:4735,0) rectangle (axis cs:4777.7,6.8883864521208e-05);
\draw[draw=none,fill=red,fill opacity=0.5] (axis cs:4777.7,0) rectangle (axis cs:4820.4,9.08480897597127e-05);
\draw[draw=none,fill=red,fill opacity=0.5] (axis cs:4820.4,0) rectangle (axis cs:4863.1,0.000103134212374834);
\draw[draw=none,fill=red,fill opacity=0.5] (axis cs:4863.1,0) rectangle (axis cs:4905.8,0.000124517504250251);
\draw[draw=none,fill=red,fill opacity=0.5] (axis cs:4905.8,0) rectangle (axis cs:4948.5,0.000137001391414511);
\draw[draw=none,fill=red,fill opacity=0.5] (axis cs:4948.5,0) rectangle (axis cs:4991.2,0.000275881546045822);
\draw[draw=none,fill=red,fill opacity=0.5] (axis cs:4991.2,0) rectangle (axis cs:5033.9,0.000148138007587776);
\draw[draw=none,fill=red,fill opacity=0.5] (axis cs:5033.9,0) rectangle (axis cs:5076.6,0.000181300650421148);
\draw[draw=none,fill=red,fill opacity=0.5] (axis cs:5076.6,0) rectangle (axis cs:5119.3,0.000195230690850974);
\draw[draw=none,fill=red,fill opacity=0.5] (axis cs:5119.3,0) rectangle (axis cs:5162,0.000244498784154677);
\draw[draw=none,fill=red,fill opacity=0.5] (axis cs:5162,0) rectangle (axis cs:5204.7,0.000265128098686509);
\draw[draw=none,fill=red,fill opacity=0.5] (axis cs:5204.7,0) rectangle (axis cs:5247.4,0.000267748478962571);
\draw[draw=none,fill=red,fill opacity=0.5] (axis cs:5247.4,0) rectangle (axis cs:5290.1,0.000298995277726177);
\draw[draw=none,fill=red,fill opacity=0.5] (axis cs:5290.1,0) rectangle (axis cs:5332.8,0.000382587880589441);
\draw[draw=none,fill=red,fill opacity=0.5] (axis cs:5332.8,0) rectangle (axis cs:5375.5,0.000902795166809445);
\draw[draw=none,fill=red,fill opacity=0.5] (axis cs:5375.5,0) rectangle (axis cs:5418.2,0.000709220754057491);
\draw[draw=none,fill=red,fill opacity=0.5] (axis cs:5418.2,0) rectangle (axis cs:5460.9,0.000987413673271349);
\draw[draw=none,fill=red,fill opacity=0.5] (axis cs:5460.9,0) rectangle (axis cs:5503.6,0.00147387738329486);
\draw[draw=none,fill=red,fill opacity=0.5] (axis cs:5503.6,0) rectangle (axis cs:5546.3,0.00272929910149955);
\draw[draw=none,fill=red,fill opacity=0.5] (axis cs:5546.3,0) rectangle (axis cs:5589,0.00386616097249671);
\draw[draw=none,fill=red,fill opacity=0.5] (axis cs:5589,0) rectangle (axis cs:5631.7,0.00361879460237947);
\draw[draw=none,fill=red,fill opacity=0.5] (axis cs:5631.7,0) rectangle (axis cs:5674.4,0.00275010146001193);
\draw[draw=none,fill=red,fill opacity=0.5] (axis cs:5674.4,0) rectangle (axis cs:5717.1,0.00152519728379585);
\draw[draw=none,fill=red,fill opacity=0.5] (axis cs:5717.1,0) rectangle (axis cs:5759.8,0.000192251862329602);
\draw[draw=none,fill=red,fill opacity=0.5] (axis cs:5759.8,0) rectangle (axis cs:5802.5,0.000112379705046981);
\draw[draw=none,fill=red,fill opacity=0.5] (axis cs:5802.5,0) rectangle (axis cs:5845.2,9.28133749667576e-05);
\draw[draw=none,fill=red,fill opacity=0.5] (axis cs:5845.2,0) rectangle (axis cs:5887.9,6.83894531483675e-05);
\draw[draw=none,fill=red,fill opacity=0.5] (axis cs:5887.9,0) rectangle (axis cs:5930.6,7.56325797604994e-05);
\draw[draw=none,fill=red,fill opacity=0.5] (axis cs:5930.6,0) rectangle (axis cs:5973.3,9.71271141948058e-05);
\draw[draw=none,fill=red,fill opacity=0.5] (axis cs:5973.3,0) rectangle (axis cs:6016,9.51865495564011e-05);
\draw[draw=none,fill=red,fill opacity=0.5] (axis cs:6016,0) rectangle (axis cs:6058.7,8.87715619937746e-05);
\draw[draw=none,fill=red,fill opacity=0.5] (axis cs:6058.7,0) rectangle (axis cs:6101.4,0.000108486215485852);
\draw[draw=none,fill=red,fill opacity=0.5] (axis cs:6101.4,0) rectangle (axis cs:6144.1,3.78595508753734e-05);
\draw[draw=none,fill=red,fill opacity=0.5] (axis cs:6144.1,0) rectangle (axis cs:6186.8,6.72275864221874e-05);
\draw[draw=none,fill=red,fill opacity=0.5] (axis cs:6186.8,0) rectangle (axis cs:6229.5,9.44572927814591e-05);
\draw[draw=none,fill=red,fill opacity=0.5] (axis cs:6229.5,0) rectangle (axis cs:6272.2,9.14908245444053e-05);
\draw[draw=none,fill=red,fill opacity=0.5] (axis cs:6272.2,0) rectangle (axis cs:6314.9,7.5237050662229e-05);
\draw[draw=none,fill=red,fill opacity=0.5] (axis cs:6314.9,0) rectangle (axis cs:6357.6,4.41385753104663e-05);
\draw[draw=none,fill=red,fill opacity=0.5] (axis cs:6357.6,0) rectangle (axis cs:6400.3,2.77241177321137e-05);
\draw[draw=none,fill=red,fill opacity=0.5] (axis cs:6400.3,0) rectangle (axis cs:6443,4.99355486570394e-06);
\end{axis}

\end{tikzpicture}

%% file: IEEEtran5/power-distribution-xavier.tex
\begin{tikzpicture}

\definecolor{darkgray176}{RGB}{176,176,176}
\definecolor{green01270}{RGB}{0,127,0}

\begin{axis}[
tick align=outside,
tick pos=left,
x grid style={darkgray176},
xlabel={Power (mW)},
xmin=3190.8, xmax=7419.2,
xtick style={color=black},
y grid style={darkgray176},
ylabel={Frequency},
ymin=0, ymax=0.00157564655841537,
ytick style={color=black}
]
\draw[draw=none,fill=green01270,fill opacity=0.5] (axis cs:3383,0) rectangle (axis cs:3421.44,3.33841105655034e-06);
\addlegendimage{ybar,ybar legend,draw=none,fill=green01270,fill opacity=0.5}
\addlegendentry{Xavier}

\draw[draw=none,fill=green01270,fill opacity=0.5] (axis cs:3421.44,0) rectangle (axis cs:3459.88,1.16844386979262e-05);
\draw[draw=none,fill=green01270,fill opacity=0.5] (axis cs:3459.88,0) rectangle (axis cs:3498.32,3.50533160937786e-05);
\draw[draw=none,fill=green01270,fill opacity=0.5] (axis cs:3498.32,0) rectangle (axis cs:3536.76,3.00456995089531e-05);
\draw[draw=none,fill=green01270,fill opacity=0.5] (axis cs:3536.76,0) rectangle (axis cs:3575.2,4.3399343735155e-05);
\draw[draw=none,fill=green01270,fill opacity=0.5] (axis cs:3575.2,0) rectangle (axis cs:3613.64,2.16996718675772e-05);
\draw[draw=none,fill=green01270,fill opacity=0.5] (axis cs:3613.64,0) rectangle (axis cs:3652.08,2.50380829241276e-05);
\draw[draw=none,fill=green01270,fill opacity=0.5] (axis cs:3652.08,0) rectangle (axis cs:3690.52,2.83764939806779e-05);
\draw[draw=none,fill=green01270,fill opacity=0.5] (axis cs:3690.52,0) rectangle (axis cs:3728.96,1.83612608110269e-05);
\draw[draw=none,fill=green01270,fill opacity=0.5] (axis cs:3728.96,0) rectangle (axis cs:3767.4,1.83612608110269e-05);
\draw[draw=none,fill=green01270,fill opacity=0.5] (axis cs:3767.4,0) rectangle (axis cs:3805.84,2.33688773958524e-05);
\draw[draw=none,fill=green01270,fill opacity=0.5] (axis cs:3805.84,0) rectangle (axis cs:3844.28,2.00304663393023e-05);
\draw[draw=none,fill=green01270,fill opacity=0.5] (axis cs:3844.28,0) rectangle (axis cs:3882.72,3.00456995089531e-05);
\draw[draw=none,fill=green01270,fill opacity=0.5] (axis cs:3882.72,0) rectangle (axis cs:3921.16,3.33841105655034e-05);
\draw[draw=none,fill=green01270,fill opacity=0.5] (axis cs:3921.16,0) rectangle (axis cs:3959.6,5.50837824330806e-05);
\draw[draw=none,fill=green01270,fill opacity=0.5] (axis cs:3959.6,0) rectangle (axis cs:3998.04,4.00609326786041e-05);
\draw[draw=none,fill=green01270,fill opacity=0.5] (axis cs:3998.04,0) rectangle (axis cs:4036.48,2.83764939806779e-05);
\draw[draw=none,fill=green01270,fill opacity=0.5] (axis cs:4036.48,0) rectangle (axis cs:4074.92,2.16996718675772e-05);
\draw[draw=none,fill=green01270,fill opacity=0.5] (axis cs:4074.92,0) rectangle (axis cs:4113.36,2.50380829241279e-05);
\draw[draw=none,fill=green01270,fill opacity=0.5] (axis cs:4113.36,0) rectangle (axis cs:4151.8,3.50533160937782e-05);
\draw[draw=none,fill=green01270,fill opacity=0.5] (axis cs:4151.8,0) rectangle (axis cs:4190.24,3.83917271503294e-05);
\draw[draw=none,fill=green01270,fill opacity=0.5] (axis cs:4190.24,0) rectangle (axis cs:4228.68,4.33993437351539e-05);
\draw[draw=none,fill=green01270,fill opacity=0.5] (axis cs:4228.68,0) rectangle (axis cs:4267.12,5.17453713765309e-05);
\draw[draw=none,fill=green01270,fill opacity=0.5] (axis cs:4267.12,0) rectangle (axis cs:4305.56,3.17149050372286e-05);
\draw[draw=none,fill=green01270,fill opacity=0.5] (axis cs:4305.56,0) rectangle (axis cs:4344,3.3384110565503e-05);
\draw[draw=none,fill=green01270,fill opacity=0.5] (axis cs:4344,0) rectangle (axis cs:4382.44,1.83612608110271e-05);
\draw[draw=none,fill=green01270,fill opacity=0.5] (axis cs:4382.44,0) rectangle (axis cs:4420.88,3.17149050372279e-05);
\draw[draw=none,fill=green01270,fill opacity=0.5] (axis cs:4420.88,0) rectangle (axis cs:4459.32,2.33688773958527e-05);
\draw[draw=none,fill=green01270,fill opacity=0.5] (axis cs:4459.32,0) rectangle (axis cs:4497.76,4.17301382068788e-05);
\draw[draw=none,fill=green01270,fill opacity=0.5] (axis cs:4497.76,0) rectangle (axis cs:4536.2,3.00456995089534e-05);
\draw[draw=none,fill=green01270,fill opacity=0.5] (axis cs:4536.2,0) rectangle (axis cs:4574.64,3.5053316093779e-05);
\draw[draw=none,fill=green01270,fill opacity=0.5] (axis cs:4574.64,0) rectangle (axis cs:4613.08,4.50685492634291e-05);
\draw[draw=none,fill=green01270,fill opacity=0.5] (axis cs:4613.08,0) rectangle (axis cs:4651.52,5.00761658482545e-05);
\draw[draw=none,fill=green01270,fill opacity=0.5] (axis cs:4651.52,0) rectangle (axis cs:4689.96,4.00609326786046e-05);
\draw[draw=none,fill=green01270,fill opacity=0.5] (axis cs:4689.96,0) rectangle (axis cs:4728.4,1.66920552827519e-05);
\draw[draw=none,fill=green01270,fill opacity=0.5] (axis cs:4728.4,0) rectangle (axis cs:4766.84,4.67377547917042e-05);
\draw[draw=none,fill=green01270,fill opacity=0.5] (axis cs:4766.84,0) rectangle (axis cs:4805.28,6.17606045461821e-05);
\draw[draw=none,fill=green01270,fill opacity=0.5] (axis cs:4805.28,0) rectangle (axis cs:4843.72,4.00609326786046e-05);
\draw[draw=none,fill=green01270,fill opacity=0.5] (axis cs:4843.72,0) rectangle (axis cs:4882.16,3.83917271503285e-05);
\draw[draw=none,fill=green01270,fill opacity=0.5] (axis cs:4882.16,0) rectangle (axis cs:4920.6,4.17301382068788e-05);
\draw[draw=none,fill=green01270,fill opacity=0.5] (axis cs:4920.6,0) rectangle (axis cs:4959.04,6.50990156027324e-05);
\draw[draw=none,fill=green01270,fill opacity=0.5] (axis cs:4959.04,0) rectangle (axis cs:4997.48,5.34145769048061e-05);
\draw[draw=none,fill=green01270,fill opacity=0.5] (axis cs:4997.48,0) rectangle (axis cs:5035.92,3.83917271503285e-05);
\draw[draw=none,fill=green01270,fill opacity=0.5] (axis cs:5035.92,0) rectangle (axis cs:5074.36,4.3399343735155e-05);
\draw[draw=none,fill=green01270,fill opacity=0.5] (axis cs:5074.36,0) rectangle (axis cs:5112.8,5.508378243308e-05);
\draw[draw=none,fill=green01270,fill opacity=0.5] (axis cs:5112.8,0) rectangle (axis cs:5151.24,7.8452659828934e-05);
\draw[draw=none,fill=green01270,fill opacity=0.5] (axis cs:5151.24,0) rectangle (axis cs:5189.68,5.84221934896303e-05);
\draw[draw=none,fill=green01270,fill opacity=0.5] (axis cs:5189.68,0) rectangle (axis cs:5228.12,7.34450432441084e-05);
\draw[draw=none,fill=green01270,fill opacity=0.5] (axis cs:5228.12,0) rectangle (axis cs:5266.56,7.0106632187558e-05);
\draw[draw=none,fill=green01270,fill opacity=0.5] (axis cs:5266.56,0) rectangle (axis cs:5305,7.17758377158315e-05);
\draw[draw=none,fill=green01270,fill opacity=0.5] (axis cs:5305,0) rectangle (axis cs:5343.44,9.34755095834107e-05);
\draw[draw=none,fill=green01270,fill opacity=0.5] (axis cs:5343.44,0) rectangle (axis cs:5381.88,8.8467892998583e-05);
\draw[draw=none,fill=green01270,fill opacity=0.5] (axis cs:5381.88,0) rectangle (axis cs:5420.32,7.17758377158332e-05);
\draw[draw=none,fill=green01270,fill opacity=0.5] (axis cs:5420.32,0) rectangle (axis cs:5458.76,0.000208650691034394);
\draw[draw=none,fill=green01270,fill opacity=0.5] (axis cs:5458.76,0) rectangle (axis cs:5497.2,0.000190289430223372);
\draw[draw=none,fill=green01270,fill opacity=0.5] (axis cs:5497.2,0) rectangle (axis cs:5535.64,0.000193627841279922);
\draw[draw=none,fill=green01270,fill opacity=0.5] (axis cs:5535.64,0) rectangle (axis cs:5574.08,0.000196966252336468);
\draw[draw=none,fill=green01270,fill opacity=0.5] (axis cs:5574.08,0) rectangle (axis cs:5612.52,0.000248711623712998);
\draw[draw=none,fill=green01270,fill opacity=0.5] (axis cs:5612.52,0) rectangle (axis cs:5650.96,0.000385586477031569);
\draw[draw=none,fill=green01270,fill opacity=0.5] (axis cs:5650.96,0) rectangle (axis cs:5689.4,0.000550837824330813);
\draw[draw=none,fill=green01270,fill opacity=0.5] (axis cs:5689.4,0) rectangle (axis cs:5727.84,0.000986500467210614);
\draw[draw=none,fill=green01270,fill opacity=0.5] (axis cs:5727.84,0) rectangle (axis cs:5766.28,0.00103156901647407);
\draw[draw=none,fill=green01270,fill opacity=0.5] (axis cs:5766.28,0) rectangle (axis cs:5804.72,0.00111169088183128);
\draw[draw=none,fill=green01270,fill opacity=0.5] (axis cs:5804.72,0) rectangle (axis cs:5843.16,0.00149393894780626);
\draw[draw=none,fill=green01270,fill opacity=0.5] (axis cs:5843.16,0) rectangle (axis cs:5881.6,0.0013370336281484);
\draw[draw=none,fill=green01270,fill opacity=0.5] (axis cs:5881.6,0) rectangle (axis cs:5920.04,0.0015006157699194);
\draw[draw=none,fill=green01270,fill opacity=0.5] (axis cs:5920.04,0) rectangle (axis cs:5958.48,0.000622613662046646);
\draw[draw=none,fill=green01270,fill opacity=0.5] (axis cs:5958.48,0) rectangle (axis cs:5996.92,0.00125357335173464);
\draw[draw=none,fill=green01270,fill opacity=0.5] (axis cs:5996.92,0) rectangle (axis cs:6035.36,0.0013186723673374);
\draw[draw=none,fill=green01270,fill opacity=0.5] (axis cs:6035.36,0) rectangle (axis cs:6073.8,0.0010966680320768);
\draw[draw=none,fill=green01270,fill opacity=0.5] (axis cs:6073.8,0) rectangle (axis cs:6112.24,0.000689381883177637);
\draw[draw=none,fill=green01270,fill opacity=0.5] (axis cs:6112.24,0) rectangle (axis cs:6150.68,0.00103824583858714);
\draw[draw=none,fill=green01270,fill opacity=0.5] (axis cs:6150.68,0) rectangle (axis cs:6189.12,0.000944770329003758);
\draw[draw=none,fill=green01270,fill opacity=0.5] (axis cs:6189.12,0) rectangle (axis cs:6227.56,0.000837941175194146);
\draw[draw=none,fill=green01270,fill opacity=0.5] (axis cs:6227.56,0) rectangle (axis cs:6266,0.000762826926421744);
\draw[draw=none,fill=green01270,fill opacity=0.5] (axis cs:6266,0) rectangle (axis cs:6304.44,0.000686043472121103);
\draw[draw=none,fill=green01270,fill opacity=0.5] (axis cs:6304.44,0) rectangle (axis cs:6342.88,0.000726104404799708);
\draw[draw=none,fill=green01270,fill opacity=0.5] (axis cs:6342.88,0) rectangle (axis cs:6381.32,0.000594237168065954);
\draw[draw=none,fill=green01270,fill opacity=0.5] (axis cs:6381.32,0) rectangle (axis cs:6419.76,0.00050243086401082);
\draw[draw=none,fill=green01270,fill opacity=0.5] (axis cs:6419.76,0) rectangle (axis cs:6458.2,0.000415632176540522);
\draw[draw=none,fill=green01270,fill opacity=0.5] (axis cs:6458.2,0) rectangle (axis cs:6496.64,0.000362217599635716);
\draw[draw=none,fill=green01270,fill opacity=0.5] (axis cs:6496.64,0) rectangle (axis cs:6535.08,0.000230350362901971);
\draw[draw=none,fill=green01270,fill opacity=0.5] (axis cs:6535.08,0) rectangle (axis cs:6573.52,0.000265403678995749);
\draw[draw=none,fill=green01270,fill opacity=0.5] (axis cs:6573.52,0) rectangle (axis cs:6611.96,0.000146890086488217);
\draw[draw=none,fill=green01270,fill opacity=0.5] (axis cs:6611.96,0) rectangle (axis cs:6650.4,0.000298787789561259);
\draw[draw=none,fill=green01270,fill opacity=0.5] (axis cs:6650.4,0) rectangle (axis cs:6688.84,0.000320487461428829);
\draw[draw=none,fill=green01270,fill opacity=0.5] (axis cs:6688.84,0) rectangle (axis cs:6727.28,0.000337179516711589);
\draw[draw=none,fill=green01270,fill opacity=0.5] (axis cs:6727.28,0) rectangle (axis cs:6765.72,0.000201973868921298);
\draw[draw=none,fill=green01270,fill opacity=0.5] (axis cs:6765.72,0) rectangle (axis cs:6804.16,0.00021699671867577);
\draw[draw=none,fill=green01270,fill opacity=0.5] (axis cs:6804.16,0) rectangle (axis cs:6842.6,0.000277088117693675);
\draw[draw=none,fill=green01270,fill opacity=0.5] (axis cs:6842.6,0) rectangle (axis cs:6881.04,0.000156905319657868);
\draw[draw=none,fill=green01270,fill opacity=0.5] (axis cs:6881.04,0) rectangle (axis cs:6919.48,7.51142487723836e-05);
\draw[draw=none,fill=green01270,fill opacity=0.5] (axis cs:6919.48,0) rectangle (axis cs:6957.92,1.50228497544764e-05);
\draw[draw=none,fill=green01270,fill opacity=0.5] (axis cs:6957.92,0) rectangle (axis cs:6996.36,3.33841105655038e-06);
\draw[draw=none,fill=green01270,fill opacity=0.5] (axis cs:6996.36,0) rectangle (axis cs:7034.8,3.33841105655038e-06);
\draw[draw=none,fill=green01270,fill opacity=0.5] (axis cs:7034.8,0) rectangle (axis cs:7073.24,0);
\draw[draw=none,fill=green01270,fill opacity=0.5] (axis cs:7073.24,0) rectangle (axis cs:7111.68,5.00761658482545e-06);
\draw[draw=none,fill=green01270,fill opacity=0.5] (axis cs:7111.68,0) rectangle (axis cs:7150.12,1.66920552827519e-06);
\draw[draw=none,fill=green01270,fill opacity=0.5] (axis cs:7150.12,0) rectangle (axis cs:7188.56,1.66920552827519e-06);
\draw[draw=none,fill=green01270,fill opacity=0.5] (axis cs:7188.56,0) rectangle (axis cs:7227,1.66920552827515e-06);
\end{axis}

\end{tikzpicture}